\documentclass{article}

    \PassOptionsToPackage{numbers, compress}{natbib}

    \usepackage[final]{neurips_2024_x}

\usepackage[utf8]{inputenc} 
\usepackage[T1]{fontenc}    
\usepackage{hyperref}       
\usepackage{url}            
\usepackage{booktabs}       
\usepackage{amsfonts}       
\usepackage{nicefrac}       
\usepackage{microtype}      
\usepackage{xcolor}         

\usepackage{hyperref}
\usepackage{url}

\usepackage{graphicx}
\usepackage{amsmath}
\usepackage{amssymb}

\usepackage{epsfig}
\usepackage{amssymb}
\usepackage{xcolor}

\usepackage{subcaption}

\usepackage{kotex}
\usepackage{color}
\usepackage{array}
\usepackage{hyperref}

\usepackage{bm}
\usepackage{nth}
\usepackage{setspace}

\newcolumntype{C}[1]{>{\centering\let\newline\\\arraybackslash\hspace{0pt}}m{#1}}
\newcolumntype{L}[1]{>{\let\newline\\\arraybackslash\hspace{0pt}}m{#1}}

\usepackage{rotating} 

\title{Noisy Label Classification using Label Noise Selection with Test-Time Augmentation Cross-Entropy and NoiseMix Learning}

\author{%
  Hansang Lee
  \quad
  Haeil Lee \\
  School of Electrical Engineering\\
  Korea Advanced Institute of Science and Technology\\
  Daehark 291, Yuseonggu, Daejeon 34141, Republic of Korea\\
  \texttt{\{hansanglee, haeil.lee\}@kaist.ac.kr}\\
  \AND
  Helen Hong \thanks{Corresponding author}\\
  Department of Software Convergence\\
  Seoul Women’s University\\
  Hwarangro 621, Nowongu, Seoul 01797, Republic of Korea\\
  \texttt{hlhong@swu.ac.kr}\\
  \And
  Junmo Kim\\
  School of Electrical Engineering\\
  Korea Advanced Institute of Science and Technology\\
  Daehark 291, Yuseonggu, Daejeon 34141, Republic of Korea\\
  \texttt{junmo.kim@kaist.ac.kr}\\
}

\begin{document}

\maketitle

\begin{abstract}
As the size of the dataset used in deep learning tasks increases, the noisy label problem, which is a task of making deep learning robust to the incorrectly labeled data, has become an important task.
In this paper, we propose a method of learning noisy label data using the label noise selection with test-time augmentation (TTA) cross-entropy and classifier learning with the NoiseMix method.
In the label noise selection, we propose TTA cross-entropy by measuring the cross-entropy to predict the test-time augmented training data.
In the classifier learning, we propose the NoiseMix method based on MixUp and BalancedMix methods by mixing the samples from the noisy and the clean label data.
In experiments on the ISIC-18 public skin lesion diagnosis dataset, the proposed TTA cross-entropy outperformed the conventional cross-entropy and the TTA uncertainty in detecting label noise data in the label noise selection process.
Moreover, the proposed NoiseMix not only outperformed the state-of-the-art methods in the classification performance but also showed the most robustness to the label noise in the classifier learning.
\end{abstract}    
\section{Introduction}

The noisy label problem represents the task of training a machine learner, mainly a deep neural network, on the training data, which consists of incorrectly labeled data~\citep{Karimi2020}.
As the size of the dataset used in deep learning tasks increases, the risk of mislabeled \textit{label noise} data being included in the dataset increases.
In addition, when labels are automatically generated from radiological reports to compose large-scale medical image datasets, there is a potential risk for label noise in medical image datasets~\citep{Ju2022}. 
Since the machine learner generally aims to make accurate predictions on all training data, the training data corrupted by the label noise leads to deterioration of the performance of the machine learner.
Due to the excellent memorization characteristics of the deep neural networks, the noisy label learning of deep neural networks can be especially challenging.

Several works have been proposed to improve the learning efficiency for noisy label data.
Most of them can be divided into two categories, sample-based and model-based methods.
The \textit{sample-based} method consists of the \textit{label noise selection} that finds the incorrect label data from the training data, and the \textit{classifier learning} that trains the classifier on the clean label data.
Han \textit{et al.} proposed the Co-Teaching method, which selects the label noise data by ranking the cross-entropy loss and trains two networks by teaching each other to improve the robustness~\citep{Han2018}.
Ju \textit{et al.} suggested the label noise selection with uncertainty estimation and the classifier learning with curriculum learning~\citep{Ju2022}.
The \textit{model-based} method aims to improve the robustness of the network to the noisy labels without the selection of label noise data.
Englesson \textit{et al.} proposed that the consistency regularization, e.g., MixUp~\citep{Zhang2018} and AugMix~\citep{Hendrycks2020} can enhance the robustness of the network to the label noise~\citep{Englesson2021}.
Xue \textit{et al.} suggested that self-supervised contrastive learning also can improve the network robustness to the noisy label problems~\citep{Xue2022}.

In this paper, we propose a sample-based method of noisy label learning in medical images to improve both the accuracy of the label noise selection and the robustness of the classifier learning.
To achieve this, we propose a test-time augmentation (TTA) cross-entropy for the label noise selection and a NoiseMix method for classifier learning.
In the label noise selection, we propose TTA cross-entropy by measuring the cross-entropy to predict the test-time augmented training data.
The proposed TTA cross-entropy can avoid the memorization problem of the conventional cross-entropy while improving the label noise detection performance of TTA uncertainty.
In the classifier learning, we propose the NoiseMix technique based on MixUp~\citep{Zhang2018} and BalancedMix~\citep{Galdran2021} methods by mixing the samples from the noisy and the clean label data.
By modifying the mixing rate of the label noise data in MixUp training, we can further improve the robustness of the classifier to the label noise.
We validate the effectiveness of the proposed method on the ISIC-18 public skin lesion diagnosis dataset~\citep{Codella2018}.
In experiments, the proposed TTA cross-entropy outperformed the conventional cross-entropy and the TTA uncertainty in detecting label noise data in the label noise selection process.
Moreover, the proposed NoiseMix not only outperformed the state-of-the-art methods in the classification performance but also showed the most robustness to the label noise.


\section{Methods}

\begin{figure*}[t!]
\centering
\includegraphics[width=\textwidth]{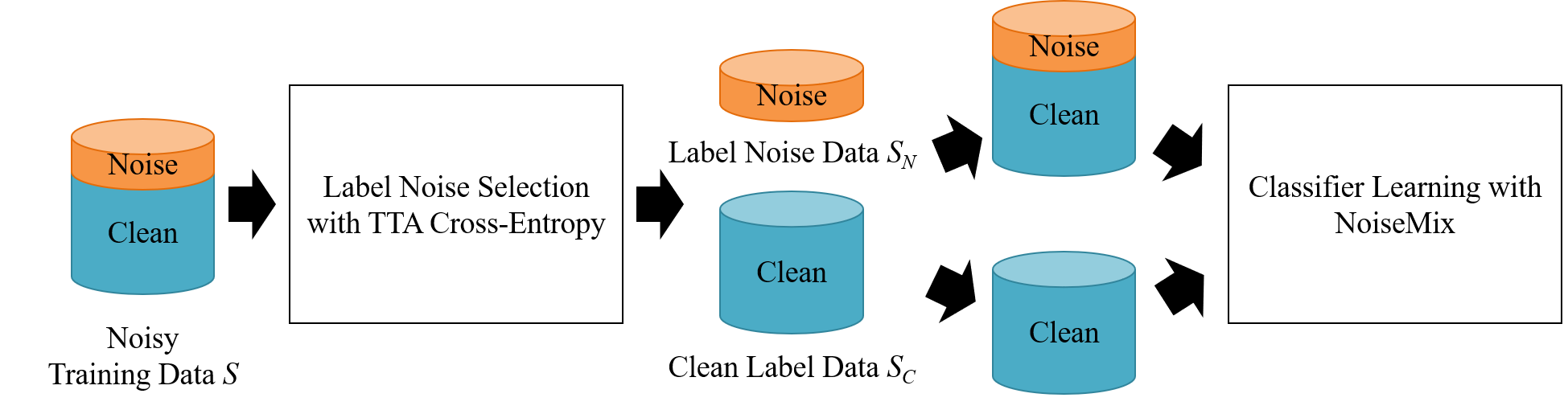} 
\caption{A pipeline of the proposed method.}
\label{fig:method}
\end{figure*}

As shown in Fig.~\ref{fig:method}, our method consists of two main steps.
First, we perform label noise selection to separate the clean and the noise label data from the noisy label data using TTA cross-entropy.
Second, we perform classifier learning on the noisy training data and the clean label data using the NoiseMix method.

\subsection{Label Noise Selection with Test-Time Augmentation Cross-Entropy}

\begin{figure*}[t!]
\centering
\includegraphics[width=\textwidth]{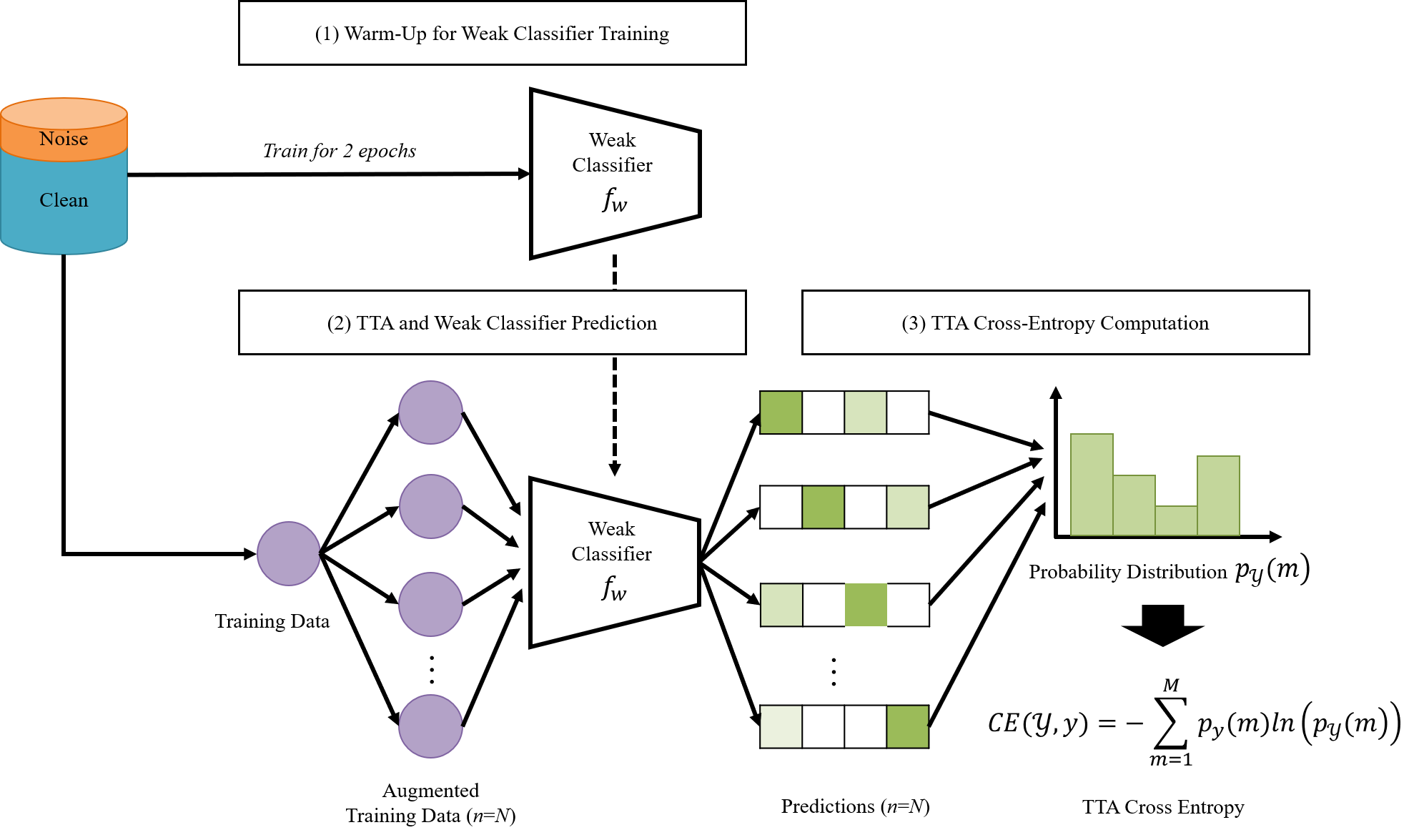} 
\caption{A pipeline of the proposed label noise selection method.}
\label{fig:method_TTACE}
\end{figure*}

In label noise selection, our aim is to separate the incorrectly-labeled \textit{label noise data} $S_{N}$ and the correctly-labeled \textit{clean label data} $S_{C}$ from the entire \textit{noisy label data} $S=\{S_{N},S_{C}\}$.
As shown in Fig.~\ref{fig:method_TTACE}, our label noise selection method consists of three steps, including (1) warm-up for weak classifier learning, (2) test-time augmentation and weak classifier prediction, and (3) TTA cross-entropy computation.

\noindent
\textbf{Warm-Up for Weak Classifier Training.}
First, we perform \textit{warm-up} that trains the \textit{weak classifier} which will provide the prediction scores of the training data for the label noise selection process.
The \textit{weak classifier} $f_{w}()$ is trained on the entire training data $S$ for a few epochs to prevent the overfitting of the classifier to the incorrect label noise.
In experiments, we perform a warm-up on the weak classifier for two epochs.

\noindent
\textbf{TTA and Weak Classifier Prediction.}
Using the trained weak classifier, we compute the prediction scores of the training data to separate the label noise and the clean label data.
In the warm-up process, the weak classifier is trained not to prevent the overfitting of the incorrect label noise, but the weak classifier still has a risk of memorizing the label noise data.
To avoid the memorization problem of a weak classifier, we perform weak classifier prediction on the augmented training data through affine transformation-based TTA instead of the training data itself.

For a training image-label pair $(x,y)\in S$, we form a set of augmented data $\mathcal{X}=\{x_{1},x_{2},...,x_{N}\}$ with affine transformation $T()$:

\begin{equation}
    x_{n} = T(x,\theta_{n})
    \label{eq:1}
\end{equation}

\noindent
where $x_{n}$ is the $n$-th augmented data and $\theta_{n}$ is the $n$-th parameter setting for affine transformation.
We then perform weak classifier prediction on these augmented data to have a set of predicted labels $\mathcal{Y}=\{y_{1},y_{2},...,y_{N}\}$ as follows:

\begin{equation}
    y_{n} = f_{w}(x_{n})
    \label{eq:2}
\end{equation}

\noindent
where $y_{n}$ is the $n$-th predicted label of the augmented training data.

\noindent
\textbf{TTA Cross-Entropy Computation.}
We compute the prediction score using the weak classifier prediction of the augmented training data and select the incorrect label noise data from the training data according to the prediction score.
As an efficient prediction score to distinguish the label noise data from the clean label data, we propose a TTA cross-entropy.

For the set of predicted labels of the augmented training data $\mathcal{Y}$ we can form a probability distribution $p_{\mathcal{Y}}$ for unique labels $m=1,2,...,M$ where $p_{\mathcal{Y}}(m)$ is the ratio of the number of $y_{n}$ with label $m$ among the $N$ predicted labels.
For the probability distribution $p_{\mathcal{Y}}$, the conventional \textit{TTA uncertainty} is computed as the entropy of the distribution:

\begin{equation}
    H(\mathcal{Y}) = -\sum_{m=1}^{M}{p_{\mathcal{Y}}(m) ln\left(p_{\mathcal{Y}}(m)\right)}.
    \label{eq:3}
\end{equation}

The TTA uncertainty is relatively robust to the memorization problem of the weak classifier compared to the conventional cross-entropy of the training data prediction~\citep{Ju2022}.
However, the TTA uncertainty only evaluates the instability of the label prediction of the augmented data, not whether the training label is correct or incorrect.
Thus, it has limitations in missing the cases with incorrect labels but relatively low label uncertainty.
To overcome the limitation, we propose a \textit{TTA cross-entropy} to reflect the correctness of training labels to the TTA uncertainty as follows:

\begin{equation}
    CE(\mathcal{Y},y) = -\sum_{m=1}^{M}{p_{y}(m) ln\left(p_{\mathcal{Y}}(m)\right)},
    \label{eq:4}
\end{equation}

\noindent
where $p_{y}$ is the probability distribution of unique labels for the training label $y$, where $p_{y}(m)=1$ if the training label $y=m$ and $p_{y}(m)=0$ if $y\neq m$.

The proposed TTA cross-entropy considers both the label instability of the weak classifier prediction of the augmented training data and the correctness of the training data labels.
Thus, the TTA cross-entropy can improve the efficiency of the label noise selection compared to the conventional training data cross-entropy and TTA uncertainty.

\subsection{Classifier Training with NoiseMix}

We re-train the classifier with the noisy label data and the clean label data obtained from the label noise selection process.
We aim to improve the learning efficiency of the clean label data learner with the label noise data while preventing overfitting of the label noise.
Inspired by the BalancedMix~\citep{Galdran2021} method for class imbalance learning, we propose the NoiseMix method for noisy label learning by combining two data sampling strategies by means of MixUp~\citep{Zhang2018}.

In NoiseMix, we form a mixed training data $(\hat{x},\hat{y})$ by mixing the clean label data $(x_{C},y_{C})\in S_{C}$ with the original training data $(x,y)\in S$ as follows:

\begin{equation}
    \hat{x} = \lambda x + (1-\lambda) x_{C}, \hat{y} = \lambda y + (1-\lambda) y_{C},
    \label{eq:5}
\end{equation}

\noindent
where $\lambda$ is the MixUp coefficient determined as $\lambda \sim \text{Beta}(\alpha,1)$. 
The NoiseMix training with $(\hat{x},\hat{y})$ in \ref{eq:5} not only enables the regularized learning of the clean label data through MixUp but also reflects the effect of re-weighting of the label noise data by mixing them only with the clean label data.


\section{Experiments and Results}

\subsection{Datasets and Experimental Details}


The proposed method was validated on the public skin lesion diagnosis dataset of ISIC-18~\citep{Tschandl2018,Combalia2019,Codella2017,Codella2018,Codella2019}~\footnote{\href{https://challenge2018.isic-archive.com/}{https://challenge2018.isic-archive.com/}}.
ISIC-18 dataset consists of 10,208 dermoscopic skin lesion images (10,015 for training and 193 for validation) for seven skin disease classes.
In this paper, we formulate a binary classification task to classify images of seven skin lesions into benign (8388 for training and 157 for validation) and malignant (1627 for training and 36 for validation,) 
The benign class includes five skin diseases of AKIEC, BKL, DF, NV, and VASC, whereas the malignant class includes two skin diseases of BCC and MEL.

To construct the noisy label dataset from the ISIC-18 dataset, we generated the \textit{instance-dependent} label noise data~\citep{Ju2022} instead of random label noise.
First, we trained a ResNet-50 on the training data for two epochs with a mini-batch size of 16.
Second, using this weak classifier, we measured the cross-entropy loss of the training data.
Third, the labels of the $r\%$ of training data with the highest losses were replaced with the labels predicted by the weak classifier, where $r$ is the ratio of the label noise data.
This instance-dependent label noise enables evaluation in a setting similar to a real noisy label environment compared to the random label noise.
Here, the ratios of the label noise data to the training data were set to $r=\{10\%,30\%,50\%\}$.

In label noise selection of the proposed method, we trained a ResNet-50 for two epochs with a mini-batch size of 16 for weak classifier training.
In TTA, a random horizontal flip, a random vertical flip, a random rotation with a degree of $-45^{\circ}\leq\theta\leq+45^{\circ}$, a random translation with a shift rate of (0.1,0.1), and a random scaling with a factor of $1\leq\sigma\leq1.2$ were applied.
In NoiseMix, the mixing hyper-parameter $\alpha$ determining the MixUp weights $\lambda\sim\text{Beta}(\alpha,1)$ was set as 0.2. 

In experiments, we evaluated (1) the effect of the proposed TTA cross-entropy for the label noise selection and (2) the effect of the proposed NoiseMix for classifier learning.
In label noise selection, we compared the proposed TTA cross-entropy with the conventional cross-entropy~\citep{Englesson2021} and the TTA uncertainty~\citep{Ju2022} by observing the ROC curve and the AUC for detecting the label noise data in the noisy training data.
In classifier learning, we compared the accuracy of the proposed NoiseMix with the results of (1) the baseline ResNet-50~\citep{He2016} trained on the noisy label data $S$, (2) the DivideMix, a state-of-the-art method for noisy label learning, trained on $S$, (3) the baseline ResNet-50 trained on the clean label data $S_{C}$ selected by the proposed TTA cross-entropy, and (4) the baseline MixUp~\citep{Zhang2018} trained on the $S_{C}$.
Both baseline ResNet-50 trained on $S$ and $S_{C}$ were trained for 100 epochs with a mini-batch size of 8.
In DivideMix, the Inception-ResNet-v2~\citep{Szegedy2017} was trained for 112 epochs with a mini-batch size of 8.
In MixUp, the ResNet-18 was trained for 100 epochs with a mini-batch size of 8 and the mixup weight parameter $\alpha$ of 0.2.

\subsection{Results}

\begin{figure*}[t!]
\centering
\begin{subfigure}{0.32\textwidth}
\includegraphics[width=\linewidth]{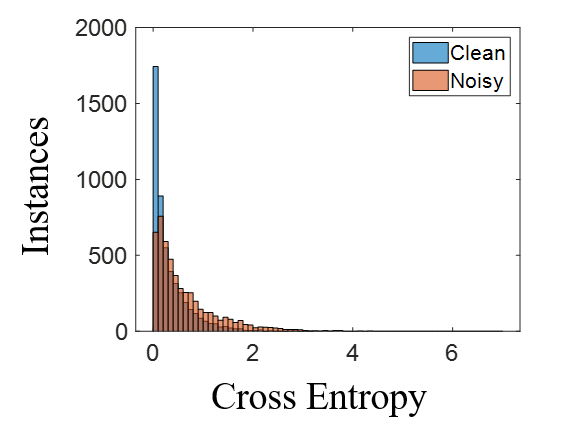} 
\caption{Cross-Entropy}
\end{subfigure}
\begin{subfigure}{0.32\textwidth}
\includegraphics[width=\linewidth]{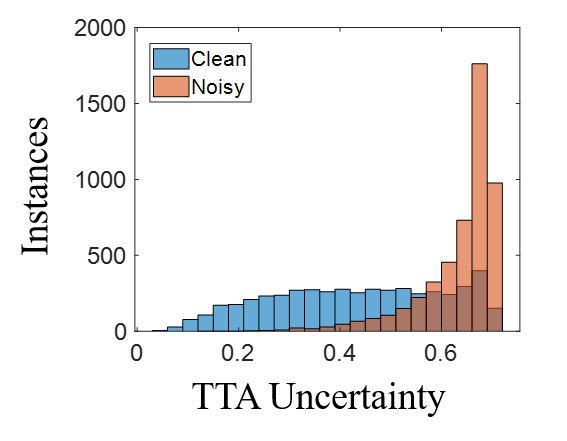}
\caption{TTA Uncertainty}
\end{subfigure}
\begin{subfigure}{0.32\textwidth}
\includegraphics[width=\linewidth]{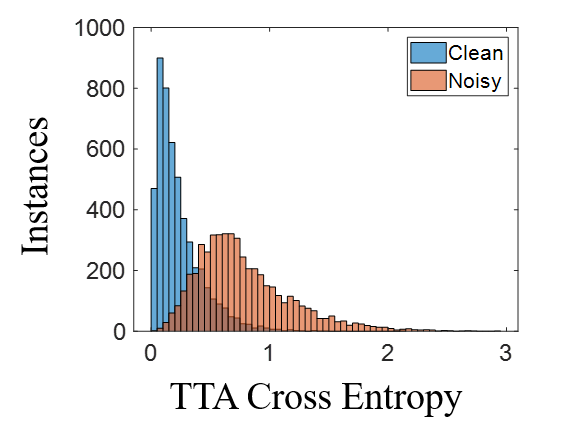}
\caption{TTA Cross-Entropy}
\end{subfigure}
\caption{Histograms of various prediction scores for the label noise (orange) and the clean label (blue) data.}
\label{Fig:Sample_Hist}
\end{figure*}

Fig.~\ref{Fig:Sample_Hist} shows the histograms of cross-entropy, TTA uncertainty, and the TTA cross-entropy for the label noise and the clean label data.
The prediction score can be considered more valuable as (1) the in-set distribution is concentrated in a limited period and (2) the distributions between two sets are far from each other, so they can be easily separated.
In Fig.~\ref{Fig:Sample_Hist} (a), The cross-entropy of the clean label data is concentrated at the low values, but the cross-entropy of the label noise data is also distributed in the low values, making separation difficult.
In Fig.~\ref{Fig:Sample_Hist} (b), The TTA uncertainty of the label noise data is concentrated at the high values, but the uncertainty of the clean label data is distributed over a wide range of periods, making incorrect detection in label noise selection.
In Fig.~\ref{Fig:Sample_Hist} (c), the distributions of TTA cross-entropy seem to be a mixture of the cross-entropy of clean label data and the TTA uncertainty of the label noise data.
Each distribution is distributed in a narrower period, and the overlap between the distributions is smaller than the cross-entropy and the TTA uncertainty.
Due to these histogram characteristics, it can be seen that the proposed TTA cross-entropy can be considered a better prediction score for distinguishing the label noise data from the clean label data compared to the cross-entropy and the TTA uncertainty.

\begin{figure*}[t!]
\centering
\begin{subfigure}{0.32\textwidth}
\includegraphics[width=\linewidth]{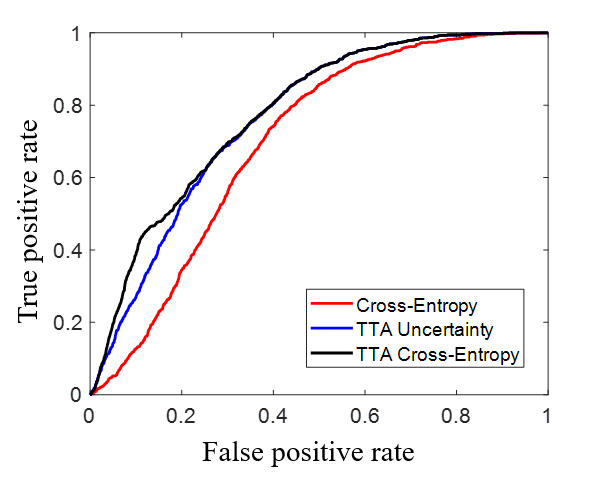} 
\caption{$r=10\%$}
\end{subfigure}
\begin{subfigure}{0.32\textwidth}
\includegraphics[width=\linewidth]{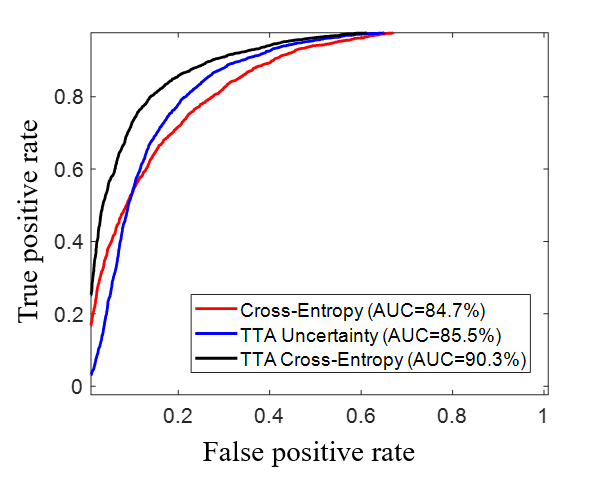}
\caption{$r=30\%$}
\end{subfigure}
\begin{subfigure}{0.32\textwidth}
\includegraphics[width=\linewidth]{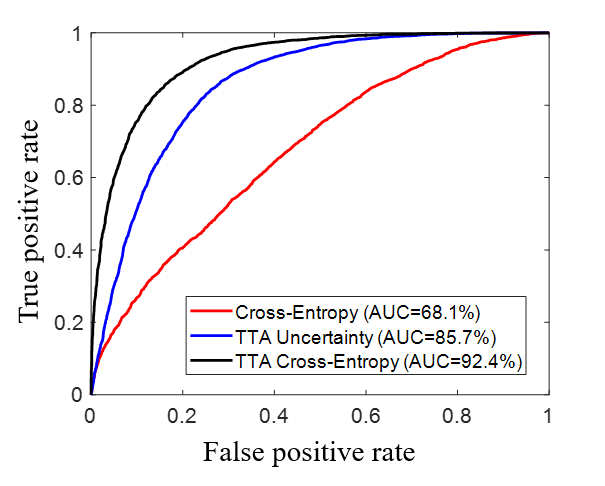}
\caption{$r=50\%$}
\end{subfigure}
\caption{ROC curves of the cross-entropy (red), TTA uncertainty (blue), and the proposed TTA cross-entropy (black) for detection of the label noise data in the label noise selection process with different label noise ratios $r$.}
\label{Fig:Sample_ROC}
\end{figure*}

\begin{table*}[b!]
\caption{Performance comparison for various methods and label noise ratios $r$.}
\centering
\begin{tabular}{L{7.5cm} | C{1.2cm} C{1.2cm} C{1.2cm}}
\hline
 & \multicolumn{3}{c}{\textbf{Label Noise Ratios}} \\
\textbf{Methods} & \textbf{10\%} & \textbf{30\%} & \textbf{50\%} \\
\hline \hline
ResNet-50~\citep{He2016} & 85.8 & 80.9 & 66.8 \\
DivideMix~\citep{Li2020} & 81.4 & 79.8 & 73.6 \\
\hline
TTA Cross-Entropy + ResNet-50 & 81.4 & 81.4 & 80.8 \\
TTA Cross-Entropy + MixUp & 85.4 & 82.5 & 75.1 \\
\textbf{TTA Cross-Entropy + NoiseMix (Ours)} & \textbf{86.7} & \textbf{83.8} & \textbf{85.5} \\
\hline
\end{tabular}
\label{tab:Acc}
\end{table*}

Fig.~\ref{Fig:Sample_ROC} shows the ROC curves and AUCs of proposed and comparative label noise selection methods for detecting the label noise data.
It can be observed that the label noise selection with cross-entropy shows low detection performance in separating label noise and clean label data.
This is because the network can be easily overfitted on the medical images due to similar appearance, and the cross-entropy is vulnerable to the memorization problem of the overfitted weak classifier.
The memorization problem seems mostly avoided by applying TTA uncertainty, and the proposed TTA cross-entropy further improves the detection performance of the label noise data regardless of label noise ratios $r$.

Table~\ref{tab:Acc} shows the accuracies of the proposed and the comparative methods in classifier learning for different label noise ratios $r$.
The performance of noisy label learning can be evaluated from two perspectives: (1) For each label noise ratio, the method with the highest accuracy can be considered as the \textit{best} noisy label learner. 
(2) For the increase in the label noise ratio, the method with the smallest decrease in performance can be considered as the \textit{most robust} noisy label learner.
From the highest accuracy perspective, it can be observed that the proposed method with NoiseMix outperformed not only the baseline ResNet-50 and MixUp but also the state-of-the-art DivideMix regardless of label noise ratios.
In the noise robustness perspective, when the label noise ratio $r$ increases from 10\% to 50\%, the MixUp and DivideMix decrease by 10.3\%p and 7.8\%p, respectively, while the performance of the proposed method decreases only 1.2\%p.
It can be confirmed that the proposed NoiseMix enables not only the best performance but also the most robust noisy label learning.

\section{Conclusions}

In this paper, we proposed a method of learning noisy label data using the label noise selection with TTA cross-entropy and the classifier learning with the NoiseMix method.
In the label noise selection, the proposed TTA cross-entropy improved the accuracy of selecting label noise data by preventing the memorization problem of the conventional cross-entropy and reflecting the label correctness to the TTA uncertainty.
In the classifier learning, the proposed NoiseMix enhanced the robustness by reflecting the effect of re-weighting the label noise data to the conventional MixUp.
As a result, our TTA cross-entropy outperformed the conventional cross-entropy and TTA uncertainty in noisy label noise selection.
Furthermore, our NoiseMix outperformed the existing training techniques without designing loss functions or weighting schemes.

\subsubsection*{Acknowledgments}
This work was supported by the National Research Foundation of Korea(NRF) grant funded by the Korea government(MSIT) (No. 2020 R1A2C1102140), and the Korea Medical Device Development Fund grant funded by the Korea government (the Ministry of Science and ICT, the Ministry of Trade, Industry and Energy, the Ministry of Health \& Welfare, the Ministry of Food and Drug Safety) (Project Number: 9991007550, KMDF\_PR\_20200901\_0269)

\bibliography{main}
\bibliographystyle{ieeetr}


\end{document}